\documentclass[letterpaper,10pt,conference]{ieeeconf}
\usepackage{graphicx,amsmath,amssymb,array,booktabs,multirow,makecell}
\ifdefined\fi
\usepackage[nocompress]{cite}
\usepackage{dblfloatfix}
\usepackage{needspace}
\graphicspath{{figures/}}

\begin{document}
\title{\LARGE \bf V3DSAM: Semantic-Geometric Loop Verification and Geometry-Anchored Persistent Instance Mapping for Feed-Forward Dense SLAM}
\author{%
\authorblockN{Ye Maolin, Zhao Dongqing\authorrefmark{1}, Li Linyang, Xue Shuai,\\
Liu Yuan, Wang Yiyuan, and Li Wen}%
\authorblockA{Information Engineering University\\[0.1ex]{\footnotesize\textsuperscript{*}Corresponding author: Zhao Dongqing.}}%
}
\maketitle
\begingroup\renewcommand{\thefootnote}{}\footnotetext{This work has been submitted to the IEEE for possible publication. Copyright may be transferred without notice, after which this version may no longer be accessible.}\endgroup
\thispagestyle{empty}
\pagestyle{empty}
\begin{abstract}
To address unreliable cross-view association in long-sequence feed-forward dense SLAM, this letter presents V3DSAM, an open-vocabulary semantic dense SLAM system that establishes physical correspondences at both the scene and object levels. The system combines feed-forward geometric estimates from VGGT-SLAM 2.0 with open-vocabulary semantic observations from SAM3. We introduce semantic-geometric dual-scale loop verification, which identifies false loop closures using pixel-level bidirectional reprojection and object-level 3D spatial correspondences, without accessing backbone-internal features. We further propose geometry-anchored persistent instance mapping, which defines physical instances through class-constrained 3D connectivity in an optimized global coordinate system and assigns global IDs without inter-frame identity propagation. The proposed verifier rejects all four false loop candidates in our TUM RGB-D evaluation, while long-sequence ScanNet experiments show improved instance identity consistency. Thus, V3DSAM reduces dependence on internal backbone features and temporal ID propagation for robust long-term cross-view association.
\end{abstract}

\noindent\textbf{\textit{Index Terms---}}
SLAM, Semantic Scene Understanding,
Loop Closure Detection, Persistent Instance Mapping.
\section{Introduction}
Visual simultaneous localization and mapping (SLAM) is fundamental to autonomous navigation, augmented reality, and 3D scene understanding. Recently, visual geometry foundation models such as VGGT have enabled feed-forward estimation of camera poses and dense scene geometry through large-scale multiview pretraining, providing a new geometric basis for dense visual SLAM \cite{r1}. Meanwhile, open-vocabulary vision models, including the SAM family and CLIP, enable flexible category- and object-level semantics in 3D scene representations, facilitating semantic SLAM systems that combine geometric reconstruction with open-vocabulary scene understanding \cite{r2,r3,r4,r5}.

\begin{figure}[!t]
\centering
\includegraphics[width=0.98\columnwidth]{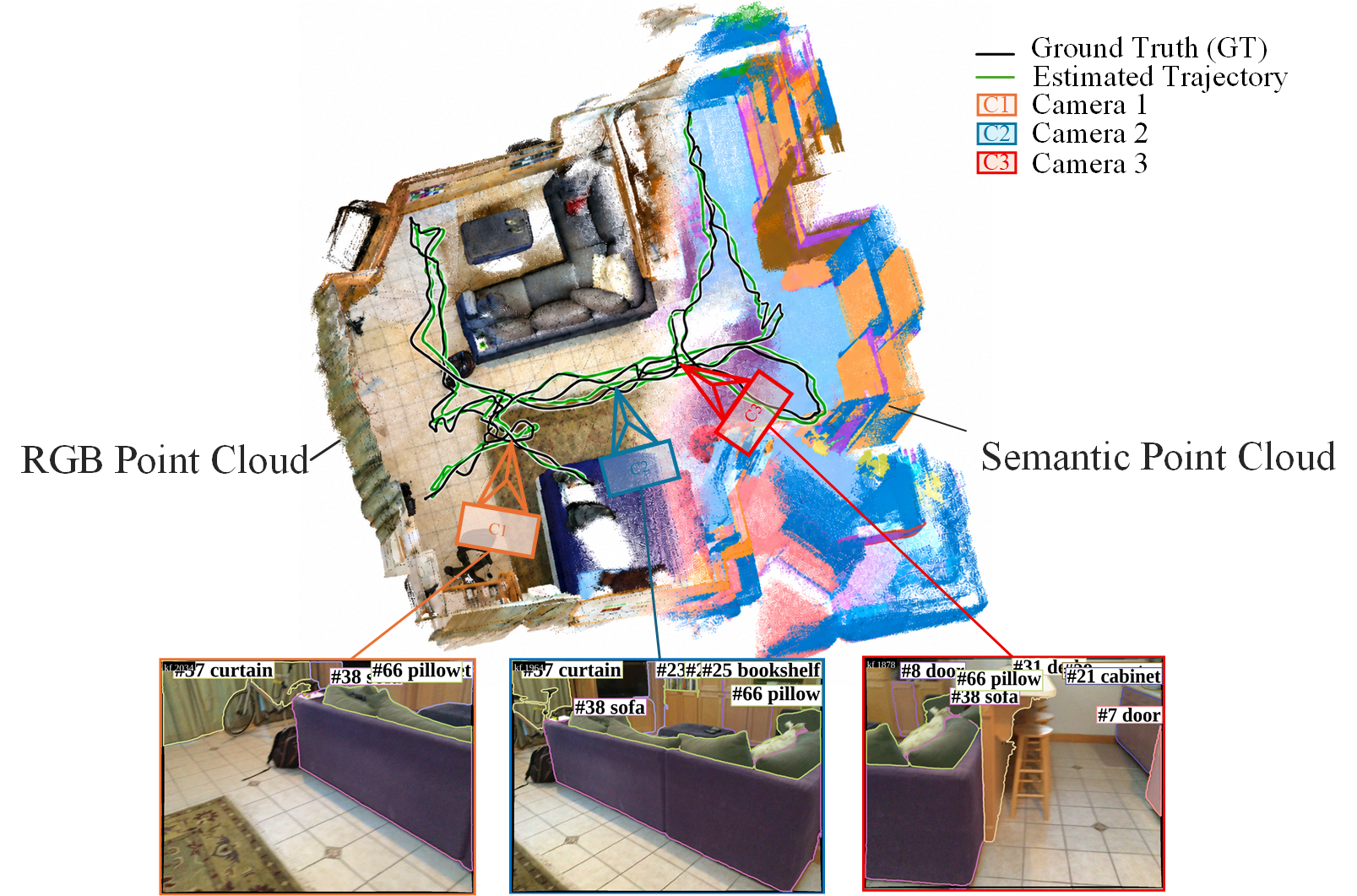}
\caption{Dense RGB and semantic point clouds, camera trajectories, and instance-labeled observations from three viewpoints (C1--C3).}
\label{fig:overview}
\end{figure}

A central challenge in extending local feed-forward geometry to long-sequence SLAM is establishing reliable and persistent physical correspondences across observations separated in time and viewpoint. The challenge manifests as scene-level false loop closures and object-level instance drift. At the scene level, appearance-based image retrieval can efficiently generate revisit candidates, but visual similarity does not guarantee that a physical location has been revisited; false-positive loop constraints may cause trajectory drift and map deformation. At the object level, association based on local temporal propagation or 2D appearance features may suffer from identity fragmentation and ID reassignment under long temporal gaps and large viewpoint changes.

Fig.~1 visualizes the outputs of V3DSAM, our open-vocabulary semantic dense SLAM system. Its main contributions are:

1) \emph{Semantic-geometric dual-scale loop verification.} Using only explicit feed-forward geometry and external open-vocabulary semantics, the method distinguishes true and false loops through bidirectional pixel-level reprojection and object-level 3D spatial correspondence. It does not rely on backbone-internal features and suppresses trajectory drift caused by false-positive loops.

2) \emph{Geometry-anchored persistent instance mapping.} Multiview semantic observations are fused in a globally optimized world coordinate system. Voxelization and class-constrained 3D connectivity analysis form physical instances, and each valid connected component receives a globally unique instance ID. This avoids inter-frame temporal ID propagation and improves identity consistency across long-term revisits.

\section{Related Work}
\subsection{Feed-Forward Visual Geometry and Dense SLAM}
Traditional iterative visual SLAM systems, such as ORB-SLAM3 and DROID-SLAM, estimate camera poses and 3D structure by iterative bundle adjustment with hand-crafted features or dense photometric constraints \cite{r8,r9,r10}. Recently, visual geometry foundation models, including DUSt3R, MASt3R, and VGGT, have learned global geometric priors from large-scale multiview pretraining. Given one or more images, they directly predict camera poses, dense depth, and 3D point clouds without frame-by-frame iterative optimization \cite{r1}, \cite{r11}, \cite{r12}. Building on these models, MASt3R-SLAM, VGGT-SLAM, and VGGT-SLAM 2.0 combine keyframe management, local maps, loop closure, and global optimization to extend feed-forward geometry to long-sequence dense SLAM \cite{r6}, \cite{r7}, \cite{r13}. This work further investigates the consistency of cross-view association in long sequences.

\subsection{Loop Candidate Retrieval and Verification}
Loop-closure processing comprises candidate retrieval (loop detection) and loop verification. The former retrieves potential revisit pairs from historical keyframes, whereas the latter rejects false candidates to prevent erroneous constraints from affecting global optimization.

Candidate retrieval commonly uses global image descriptors such as NetVLAD and SALAD \cite{r14}, \cite{r15}; Hydra instead employs hierarchical 3D scene structures for loop association \cite{r16}. Conventional verification often relies on geometric registration such as ICP, which performs reliably in scale-consistent iterative SLAM but may be challenged by noise in feed-forward geometry and semantic ambiguity \cite{r17}.

In feed-forward SLAM, MASt3R-SLAM reprocesses candidate frames through its backbone for two-view inference and verifies them using matching quality, coupling verification to the encoder and introducing additional inference cost \cite{r13}. EC3R-SLAM uses a homography-RANSAC inlier ratio for loop verification, relying on geometric cues without object-level semantics \cite{r18}. VGGT-SLAM 2.0 extracts the signal $\alpha_{match}$ from VGGT attention layers to verify retrieved candidates without additional training; however, this signal is tied to a specific network architecture and is less readily reusable. Establishing interpretable cross-view physical correspondence from explicit feed-forward 3D geometry and independent semantic observations, without using backbone-specific features, therefore remains of interest.

\subsection{Cross-View Instance Association and Open-Vocabulary Instance Mapping}
Early instance SLAM methods often depend on predefined semantic categories or local instance association. MaskFusion establishes instance correspondences using 2D masks and temporal tracking \cite{r19}, PanopticFusion fuses panoptic observations into a 3D voxel map to maintain identities \cite{r20}, and SGS-SLAM jointly optimizes geometry and semantics with 3D Gaussian representations \cite{r21}. Nevertheless, stable identities remain challenging across long temporal gaps and large viewpoint changes.

With advances in vision-language pretraining, CLIP-Fields aggregates CLIP features into 3D point clouds for point-level open-vocabulary queries, primarily targeting offline multiview fusion \cite{r22}. OVO combines semantic categories, vision-language embeddings, temporal tracking, and spatial constraints to build instance-level maps, but long-term identity persistence remains limited \cite{r23}. These approaches place greater emphasis on semantic representation richness than on long-term identity consistency.

For semantic mapping built on feed-forward geometry, SceneVGGT back-projects SAM masks to construct semantic point clouds, but instance partitioning still depends heavily on inter-frame 2D semantic-label matching \cite{r24}. IRIS-SLAM fine-tunes the feed-forward backbone to learn dedicated instance embeddings, coupling its approach to internal model representations \cite{r25}. In multiview 3D instance construction, Open3DIS and MaskClustering use 2D masks to guide 3D instance aggregation, but do not specifically address noise in predicted feed-forward geometry \cite{r26}, \cite{r27}. Maintaining long-term global instance identities under such geometric noise, without temporal ID propagation, remains an open challenge.

\section{Method}
\subsection{Overview of V3DSAM}
As shown in Fig.~2, V3DSAM takes a continuous monocular RGB sequence as input. VGGT-SLAM 2.0 and SAM3 provide camera intrinsics and extrinsics, dense depth, and semantic masks. An appearance-retrieval module generates loop candidates, which are processed by semantic-geometric dual-scale verification. Only candidates satisfying both consistency constraints enter global geometric optimization. In the resulting optimized world coordinate system, multiview semantic observations are fused by geometry-anchored instance assignment to construct a persistent dense 3D instance map.

\begin{figure*}[!t]
\centering
\includegraphics[width=0.96\textwidth]{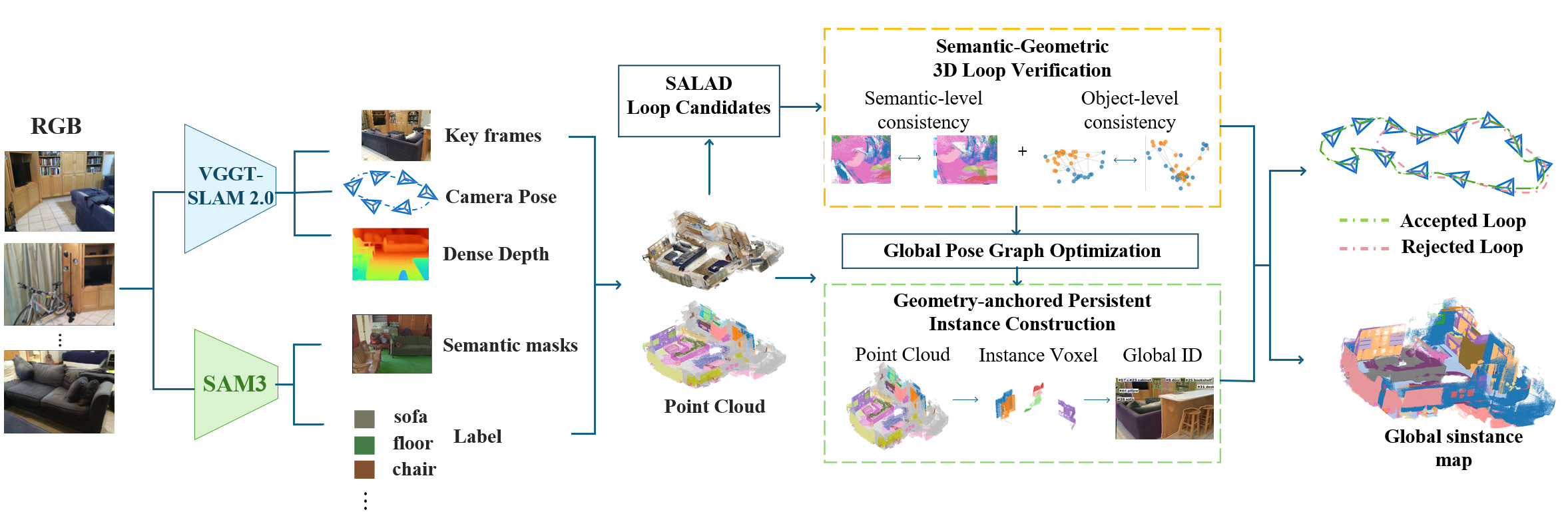}
\caption{Overall system architecture of V3DSAM.}
\label{fig:framework}
\end{figure*}
\subsection{Semantic-Geometric Dual-Scale Loop Verification}
We formulate loop-candidate verification as a binary hypothesis test. Given a query frame $q$ and a candidate frame $c$, the feed-forward geometry provides their relative transform $\mathbf{T}_{c\leftarrow q}\in SE(3)$. Hypothesis $H_1$ denotes a true loop, where the two frames observe the same physical region; $H_0$ denotes a false loop, where the views appear similar but lack spatial-semantic correspondence.

We define a semantic-geometric dual-scale criterion from pixel-level consistency $C_{pix}$ and object-level consistency $C_{obj}$. A candidate is accepted only when both scores satisfy their respective thresholds, as illustrated in Fig.~3.

\begin{figure}[!t]
\centering
\includegraphics[width=\columnwidth]{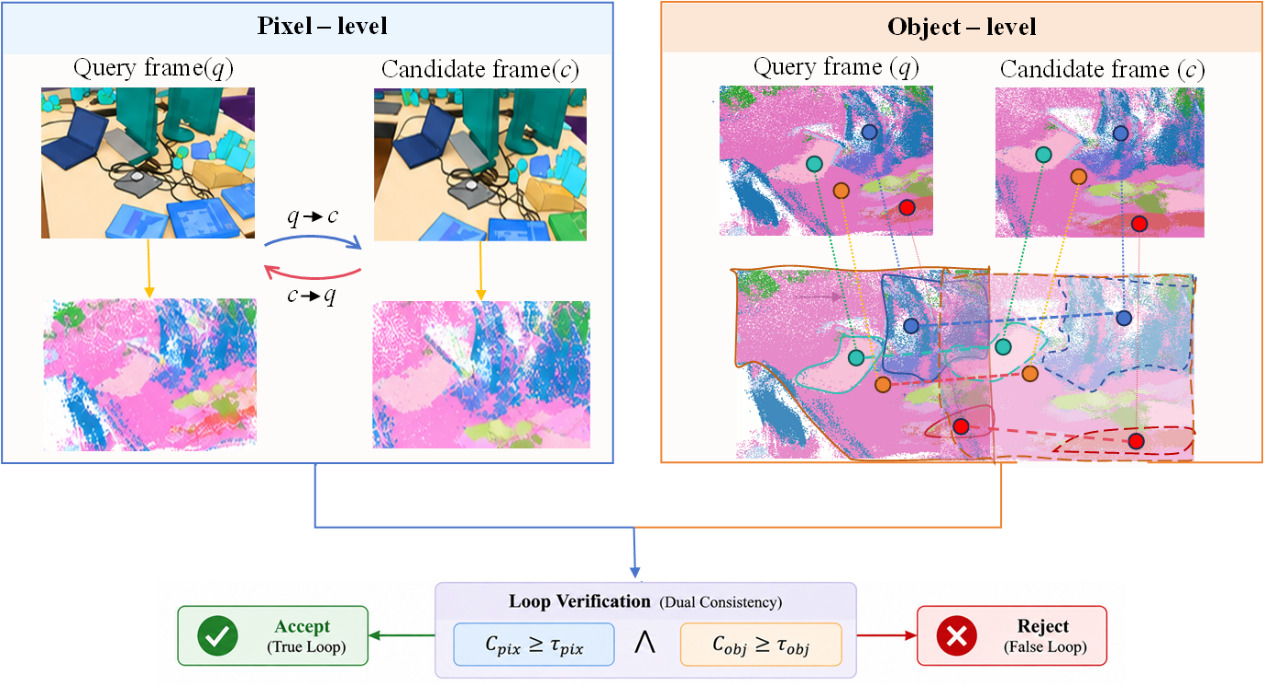}
\caption{Principle of semantic-geometric dual-scale loop verification.}
\label{fig:loop}
\end{figure}
For a valid semantic pixel at $\left( u_{i},v_{i} \right)$ in keyframe $k$, its depth and the camera intrinsics recover the corresponding point in the camera coordinate system:

\begin{equation}\label{eq:1}
\mathbf{p}_{i}^{k} = D_{k}\left( u_{i},v_{i} \right)\begin{bmatrix}
\left( u_{i} - c_{x} \right)/f_{x} \\
\left( v_{i} - c_{y} \right)/f_{y} \\
1
\end{bmatrix},\quad l_{i} = S_{k}\left( u_{i},v_{i} \right)
\tag{1}
\end{equation}

Here, $D_{k}\left( u_{i},v_{i} \right)$ is the depth at the pixel, $f_{x},f_{y},c_{x},c_{y}$ are the camera intrinsic parameters, $\mathbf{p}_{i}^{k}$ is the back-projected 3D point, and $l_{i}$ is the open-vocabulary semantic class supplied by SAM3. Applying this operation to the query and candidate frames yields two semantic 3D point sets.

Unidirectional reprojection is affected by image-boundary truncation. A reprojected point is counted only if its image coordinates lie within $u \in [ 0,W),v \in [ 0,H)$. We define the projection validity indicator as

\begin{equation}\label{eq:2}
\mathbb{I}(u,v) = \left\{ \begin{matrix}
1, & 0 \leq u < W,\ 0 \leq v < H \\
0, & \text{otherwise}
\end{matrix} \right.\ 
\tag{2}
\end{equation}

where $W$ and $H$ denote image width and height.

To alleviate statistical bias from boundary truncation, we use bidirectional reprojection for pixel-level semantic consistency. Query-frame points are transformed into the candidate-frame coordinate system by $\mathbf{T}_{c \leftarrow q}$ and perspectively projected. Swapping $q$ and $c$ repeats the operation in the reverse direction. The pixel-level consistency score is 
\begin{equation}\label{eq:3}
C_{pix} = \frac{1}{2}\left( \frac{N_{match}^{q \rightarrow c}}{N_{valid}^{q \rightarrow c}} + \frac{N_{match}^{c \rightarrow q}}{N_{valid}^{c \rightarrow q}} \right)
\tag{3}
\end{equation}

Here, $N_{valid}$ denotes the number of valid semantic reprojections in a given direction, and $N_{match}$ the number for which the source 3D point's semantic class matches the target pixel's class. A higher $C_{pix}$ indicates stronger agreement between the two views under the hypothesized relative pose.

$C_{pix}$ measures local dense semantic alignment but can be disturbed by fragmented masks and partial occlusion. We therefore introduce an object-level constraint. Within each frame, 3D semantic points of the same class are spatially clustered; noisy clusters containing fewer than $N_{\min}$ points are filtered out. Each valid cluster is represented by its 3D centroid. Object correspondence between frames is then established using the known relative pose $\mathbf{T}_{c \leftarrow q}$. A match must satisfy two conditions: the semantic classes agree, and the 3D Euclidean distance between centroids after pose transformation is below a preset threshold. Bidirectional matching is performed for $q \rightarrow c$ and $c \rightarrow q$, and the number of matched objects in each direction is counted.

\begin{equation}\label{eq:4}
C_{obj} = \frac{1}{2}\left( \frac{M_{q \rightarrow c}}{N_{q}} + \frac{M_{c \rightarrow q}}{N_{c}} \right)
\tag{4}
\end{equation}

Here, $N_{q}$ and $N_{c}$ denote the numbers of valid objects participating in 3D matching in the two views, while $M_{q \rightarrow c}$ and $M_{c \rightarrow q}$ are the numbers of successful 3D correspondences in the respective directions.

$C_{pix}$ emphasizes local dense alignment, whereas $C_{obj}$ captures the global arrangement of physical entities. Together they characterize semantic-geometric agreement from complementary perspectives. Loop verification is expressed as a binary objective with hard threshold constraints:

\begin{equation}\label{eq:5}
\mathcal{L}_{\text{loop}}\left( \mathbf{T}_{c \leftarrow q} \right) = \left\{ \begin{matrix}
0, & C_{pix} \geq \tau_{pix}\, \land \, C_{obj} \geq \tau_{obj} \\
 + \infty, & \text{otherwise}
\end{matrix} \right.\ 
\tag{5}
\end{equation}

If $\mathcal{L}_{\text{loop}} = 0$, the loop constraint is accepted for back-end global optimization. If $\mathcal{L}_{\text{loop}} = + \infty$, the candidate is treated as a false loop, $H_{0}$, and discarded without entering subsequent optimization.

\subsection{Class-Constrained Instance Construction and Geometry-Anchored Identity Maintenance}
Conventional instance association frequently propagates instance IDs through matches between adjacent frames. This makes identity depend strongly on temporal order: long-term revisits or severe occlusions can break the tracking chain, causing ID changes and identity resets.

We instead formulate instance partitioning directly in the unified world coordinate system obtained by global optimization. Multiview observations yield a noisy semantic point set $\mathcal{P} = \left\{ \left( \mathbf{p}_{i},l_{i} \right) \right\}$, where $\mathbf{p}_{i} \in \mathbb{R}^{3}$ represents a 3D coordinate and $l_{i}$ its semantic class label. The objective is to partition $\mathcal{P}$ into physical entities whose points are both semantically consistent and spatially connected in 3D.

The raw multiview point cloud contains repeated observations and sampling noise. We therefore introduce voxelization to regularize these observations. For all points $\left\{ \mathbf{p}_{m},l_{m} \right\}_{m \in \mathcal{M}_{v}}$ within voxel $v$, the voxel position is the coordinate mean and its semantic label is determined by majority voting:

\begin{equation}
\begin{aligned}
\mathbf{p}_v &=
\frac{1}{|\mathcal{M}_v|}
\sum_{m\in\mathcal{M}_v}\mathbf{p}_m,\\
\ell_v &=
\operatorname*{arg\,max}_{\ell}
\sum_{m\in\mathcal{M}_v}
\mathbb{I}(\ell_m=\ell).
\end{aligned}
\label{eq:voxel}
\tag{6}
\end{equation}

Here, $\mathcal{M}_{v}$ is the set of raw 3D points assigned to the voxel, and $\mathbb{I}( \cdot )$ denotes an indicator function. Equation (6) produces the regularized voxel set $\mathcal{V} = \left\{ \left( \mathbf{p}_{v},l_{v} \right) \right\}$.

A class-constrained 3D adjacency graph $G = \left( \mathcal{V},\mathcal{E} \right)$ is built over the regularized voxels. An edge between voxels $v_{i}$ and $v_{j}$ is created if and only if they have the same semantic class and their Euclidean distance is below neighborhood threshold $\epsilon_{c}$:

\begin{equation}\label{eq:7}
\left( v_{i},v_{j} \right)\in\mathcal{E}\Longleftrightarrow l_{i} = l_{j} \land \lVert \mathbf{p}_{i} - \mathbf{p}_{j} \rVert_{2} < \epsilon_{c}
\tag{7}
\end{equation}

Physical instances are defined by paths in the adjacency graph: two voxels of the same class belong to the same instance if a path of neighboring same-class voxels connects them. This connectivity relation is an equivalence relation and partitions the voxel set into disjoint connected components, each representing a physical 3D instance. A KD-tree performs neighborhood searches among voxels of the same class, and a union-find algorithm efficiently computes the connected components. After discarding noisy components with fewer voxels than a minimum threshold, each remaining component receives a unique global instance ID that serves as the persistent identity of the physical entity.

To associate 3D instances with 2D observations, instance voxels are projected into keyframe images, producing an instance mask $M_{3D}$. Its intersection-over-union with the SAM 2D mask $M_{2D}$ is computed under the same semantic class constraint:

\begin{equation}\label{eq:8}
IoU\left( M_{3D},M_{2D} \right) = \frac{\left| M_{3D} \cap M_{2D} \right|}{\left| M_{3D} \cup M_{2D} \right|}
\tag{8}
\end{equation}

The highest-$IoU$ 2D mask within the same class is assigned the global instance ID. If no mask reaches the minimum matching threshold, no 2D instance label is assigned in that frame.

\section{Experiments and Results}
\subsection{Experimental Setup}
\subsubsection{Datasets}
Experiments use ScanNet \cite{r28} and TUM RGB-D \cite{r29} indoor datasets. The TUM RGB-D experiments cover eight Freiburg1 sequences. With a unified front-end configuration, these sequences yield 15 SALAD loop candidates: 11 true and four false loops. Together with four ScanNet scenes, the evaluation contains 46 candidate frame pairs. All hyperparameters of the loop-verification module are selected using TUM RGB-D only and transferred directly to ScanNet without retuning, to test cross-dataset generalization under fixed parameters.

\subsubsection{Parameters and Hardware}
VGGT-SLAM 2.0 provides the feed-forward geometry, and SAM3 generates open-vocabulary semantic observations for keyframes. On TUM RGB-D, each submap contains 32 frames and the SALAD candidate threshold is 0.95.

For loop verification, the pixel- and object-level consistency thresholds are $\tau_{pix} = 0.10$ and $\tau_{obj} = 0.25$, respectively. Within a frame, same-class 3D clustering uses a 0.3 m neighborhood threshold and removes sparse clusters containing fewer than 80 points. Persistent instance mapping uses a 0.02 m voxel resolution and a 0.05 m 3D neighborhood connection distance, and filters out connected components containing fewer than 100 voxels.

Experiments are conducted on an NVIDIA RTX 4090 GPU and an Intel i7-14700 CPU.

\subsubsection{Evaluation Metrics}
Trajectory accuracy is measured using absolute trajectory error (ATE-RMSE). Loop verification is evaluated using a confusion matrix, precision, recall, and F1-score.

Following the overlap-based loop-labeling principle in \cite{r31}, we define loop ground truth from the extent of shared 3D visibility computed using ground-truth poses and depth.

Ground-truth loop labels are computed from dataset poses and depth. Symmetric geometric overlap is defined as the average of the two directions:

\begin{equation}\label{eq:9}
O^{GT}(q,c) = \frac{1}{2}\left( \frac{N_{q \rightarrow c}^{vis}}{N_{q}^{valid}} + \frac{N_{c \rightarrow q}^{vis}}{N_{c}^{valid}} \right)
\tag{9}
\end{equation}

Here, $N_{q}^{valid}$ denotes the number of valid depth points in the query frame and $N_{q \rightarrow c}^{vis}$ the number that establish depth-consistent overlap in the candidate frame after transformation using the ground-truth pose. A candidate is labeled a true loop only if its temporal separation exceeds 20 frames and its overlap is at least 0.30. Ground truth is used for offline evaluation only, not during retrieval, verification, or optimization.

Instance mapping is also evaluated using standard instance-SLAM measures: multiple object tracking accuracy (MOTA), 4D panoptic quality (4D-PQ), and ID-switch count (IDS), for comparison with prior instance mapping methods \cite{r32}, \cite{r33}. These composite measures account for detection, segmentation, misses, false positives, and identity switches, but do not fully characterize identity persistence across long temporal intervals and loop revisits. We therefore additionally report identity-discrimination consistency (IDC), long-term reidentification recall (ReID-R), and instance coverage as targeted auxiliary metrics.

For IDC, let $N_{g}$ be the number of visible observations of ground-truth instance $g$, and $N_{g,j}$ the number of these observations assigned predicted global ID $j$. Then

\begin{equation}\label{eq:11}
IDC = \frac{\sum_{g \in \mathcal{G}}^{}\max_{j}N_{g,j}}{\sum_{g \in \mathcal{G}}^{}N_{g}}\ 
\tag{10}
\end{equation}

Here, $\max_{j}N_{g,j}$ is the highest occurrence count of any predicted ID assigned to ground-truth instance $g$. Higher IDC indicates that an object retains a more stable identity throughout the sequence.

For long-term Re-ID recall, first define the set of long-term visible instances:

\begin{equation}\label{eq:12}
\mathcal{G}_{L} = \left\{ g \mid t_{g}^{last} - t_{g}^{first} \geq 0.25T \right\}
\tag{11}
\end{equation}

where $T$ is the sequence length. The recall is then computed as

\begin{equation}\label{eq:13}
ReID\text{-}R = \frac{N_{same\text{~}ID}}{N_{long\text{-}term}}
\tag{12}
\end{equation}

Here, $N_{long\text{-}term} = |\mathcal{G}_{L}|$ is the number of ground-truth instances satisfying the long-term span criterion, while $N_{same\text{~}ID}$ counts instances that are successfully matched at both the first and last observations and retain the same predicted global ID. Failure to detect either endpoint or different IDs at the endpoints is counted as a Re-ID failure.

\subsection{Loop Verification and Localization Accuracy}
Table I compares false positives (FP), precision, and F1-score for different verification strategies on the same SALAD candidate set.

\begin{table}[!t]
\vspace*{10pt}
\caption{Comparison of loop-verification signals on the same SALAD candidate set}
\label{tab:1}
\centering
\footnotesize
\setlength{\tabcolsep}{1.6pt}
\renewcommand{\arraystretch}{1.13}
\begin{tabular*}{0.98\columnwidth}{@{\extracolsep{\fill}}lccc@{}}
\toprule
Verification signal & FP $\downarrow$ & Precision $\uparrow$ & F1 $\uparrow$ \\
\midrule
SALAD retrieval & 4 & 0.733 & 0.846 \\
VGGT & 0 & 1.000 & 1.000 \\
Semantic only (Category Jaccard) & 2 & 0.846 & 0.917 \\
Geometry only (Depth consistency) & 2 & 0.846 & 0.917 \\
\textbf{Semantic-Geometric 3D (Ours)} & 0 & 1.000 & 1.000 \\
\bottomrule
\end{tabular*}
\par\vspace{2pt}\raggedright\footnotesize
\emph{Note:} FP is the number of false-positive loop candidates; $\uparrow$ and $\downarrow$ indicate preferred metric directions.
\end{table}

Table I shows that semantic-only and geometry-only verification each retain two false-positive loops, whereas the joint criterion rejects all four false candidates while preserving the 11 true loops (precision and F1-score of 1.000). The two cues are complementary: semantics suppresses geometric ambiguity, while 3D correspondence rejects scenes with similar category compositions.

\begin{figure}[!t]
\vspace*{10pt}
\centering
\includegraphics[width=\columnwidth]{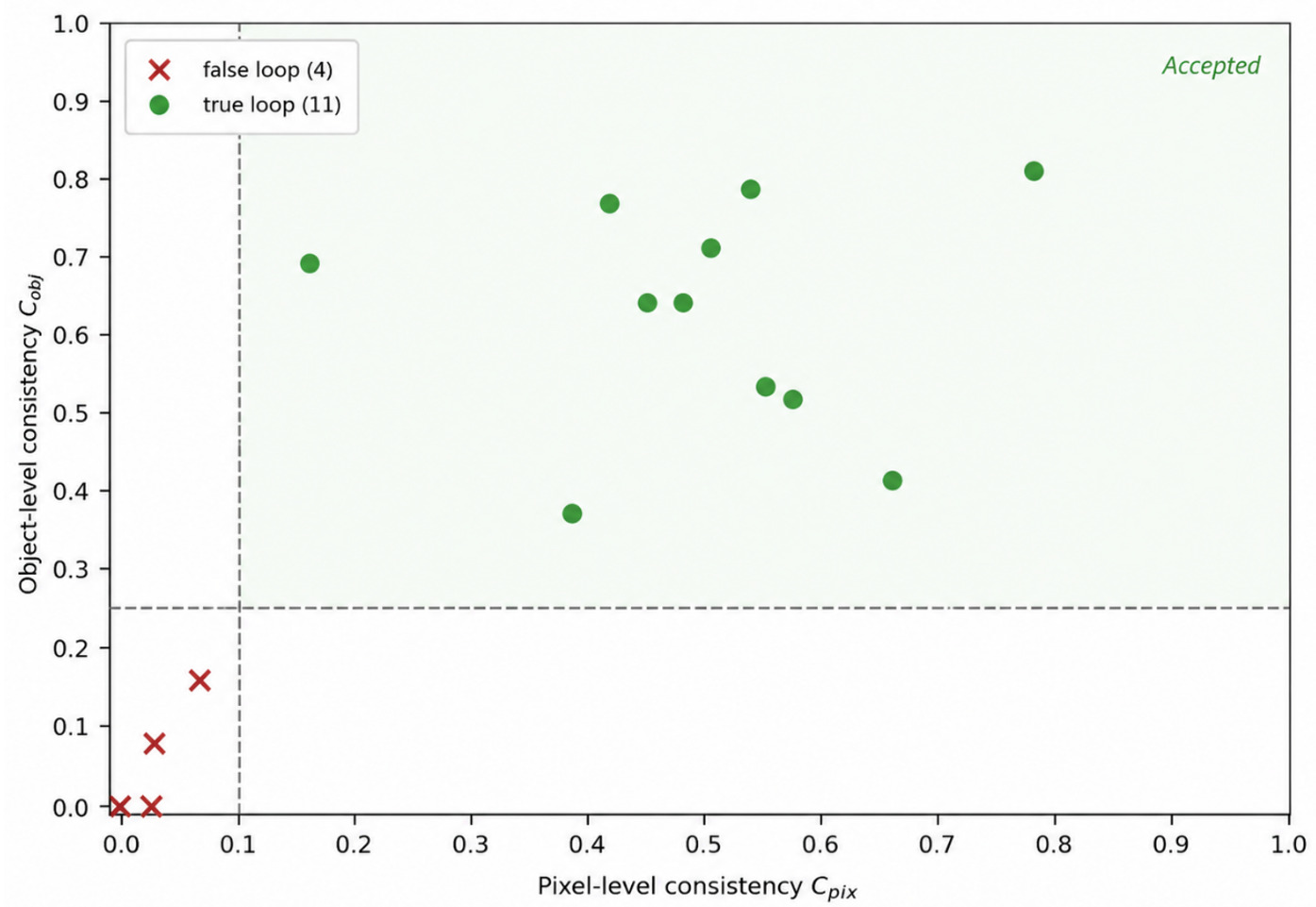}
\caption{Distribution of true and false loop candidates in pixel-level and object-level consistency scores on TUM RGB-D. Green circles indicate true loops; red crosses indicate false loops. Dashed lines mark the thresholds $\tau_{pix}=0.10$ and $\tau_{obj}=0.25$.}
\label{fig:scatter}
\end{figure}
Fig.~4 shows a clear separation between true and false loop candidates on TUM RGB-D in the two-dimensional feature space defined by pixel-level consistency $C_{pix}$ and object-level consistency $C_{obj}$. False loops often lack matchable 3D object structures that are visible in both views, resulting in low consistency scores at both the pixel and object levels. In contrast, true loops satisfy both pixel-level geometric alignment and object-level instance-matching constraints, yielding high scores on both measures.

We remove the false loops selected by the semantic-geometric dual-scale verification module before global pose optimization. Table II reports the trajectory accuracy after removing these erroneous constraints.

\begin{table*}[!t]
\vspace*{10pt}
\caption{ATE-RMSE on eight TUM RGB-D sequences}
\label{tab:2}
\centering
\footnotesize
\setlength{\tabcolsep}{4.1pt}
\renewcommand{\arraystretch}{1.10}
\begin{tabular}{@{}llcccccccc@{}}
\toprule
\textbf{Setting} & \textbf{Method} & \textbf{360} & \textbf{desk} & \textbf{desk2} & \textbf{plant} & \textbf{room} & \textbf{rpy} & \textbf{teddy} & \textbf{xyz} \\
\midrule
Calibrated & ORB-SLAM3 & - & 0.017 & 0.210 & 0.034 & - & - & - & \textbf{0.009} \\
 & DeepV2D & 0.243 & 0.166 & 0.379 & 0.203 & 0.246 & 0.105 & 0.316 & 0.064 \\
 & DeepFactors & 0.159 & 0.170 & 0.253 & 0.305 & 0.364 & 0.043 & 0.601 & 0.035 \\
 & DPV-SLAM & 0.112 & 0.018 & 0.029 & 0.021 & 0.330 & 0.030 & 0.084 & 0.010 \\
 & DPV-SLAM++ & 0.132 & 0.018 & 0.029 & 0.022 & 0.096 & 0.032 & 0.098 & 0.010 \\
 & GO-SLAM & 0.089 & \textbf{0.016} & 0.028 & 0.026 & 0.052 & \textbf{0.019} & 0.048 & 0.010 \\
 & DROID-SLAM & 0.111 & 0.018 & 0.042 & \textbf{0.016} & \textbf{0.049} & 0.026 & 0.048 & 0.012 \\
 & MASt3R-SLAM & \textbf{0.049} & \textbf{0.016} & \textbf{0.024} & 0.020 & 0.061 & 0.027 & \textbf{0.041} & \textbf{0.009} \\
\midrule
Uncalibrated & DROID-SLAM & 0.202 & 0.032 & 0.091 & 0.045 & 0.918 & 0.056 & 0.045 & \textbf{0.012} \\
 & MASt3R-SLAM & 0.070 & 0.035 & 0.055 & 0.035 & 0.118 & 0.041 & 0.114 & 0.020 \\
 & ViSTA-SLAM & 0.104 & 0.030 & 0.030 & 0.052 & 0.067 & \textbf{0.023} & 0.080 & 0.015 \\
 & VGGT-SLAM 2.0 & 0.050 & \textbf{0.025} & 0.029 & \textbf{0.026} & \textbf{0.063} & 0.026 & \textbf{0.038} & 0.014 \\
 & Ours (no verification) & 0.189$\times$ & \textbf{0.025} & \textbf{0.028} & 0.035 & 0.815$\times$ & 0.026 & 0.043 & 0.014 \\
 & \textbf{V3DSAM (Ours)} & \textbf{0.049} & \textbf{0.025} & \textbf{0.028} & 0.035 & 0.074 & 0.026 & 0.043 & 0.014 \\
\bottomrule
\end{tabular}
\par\vspace{2pt}\raggedright\footnotesize \emph{Note:} All ATE-RMSE values are reported in meters (m); lower is better. A dash denotes tracking failure; $\times$ marks drift due to false loops. \textbf{Bold} denotes the lowest ATE within each calibration setting; calibrated and uncalibrated methods are ranked separately, with ties bolded. The no-verification ablation accepts all SALAD loop candidates.
\end{table*}

The baseline results in Table II are taken from the publicly available VGGT-SLAM 2.0 study \cite{r7}. The table compares absolute trajectory error (ATE-RMSE) across eight TUM RGB-D sequences for several recent SLAM systems. "Ours (no Verification)" is an ablation that removes semantic-geometric dual-scale verification and directly accepts all candidate loops. Without verification, false loops introduce erroneous pose constraints and corrupt global trajectory optimization. As a result, severe drift and increased ATE occur on the perceptually ambiguous fr1/360 and fr1/room sequences. Enabling our verification method filters out these false candidates and prevents their constraints from degrading the back-end optimization. Overall, our method yields pose accuracy comparable to VGGT-SLAM 2.0's internal attention score $\alpha_{match}$, while suppressing false-loop interference and preserving the global correction obtained from true loops.

\begin{figure*}[!b]
\centering
\includegraphics[width=0.87\textwidth]{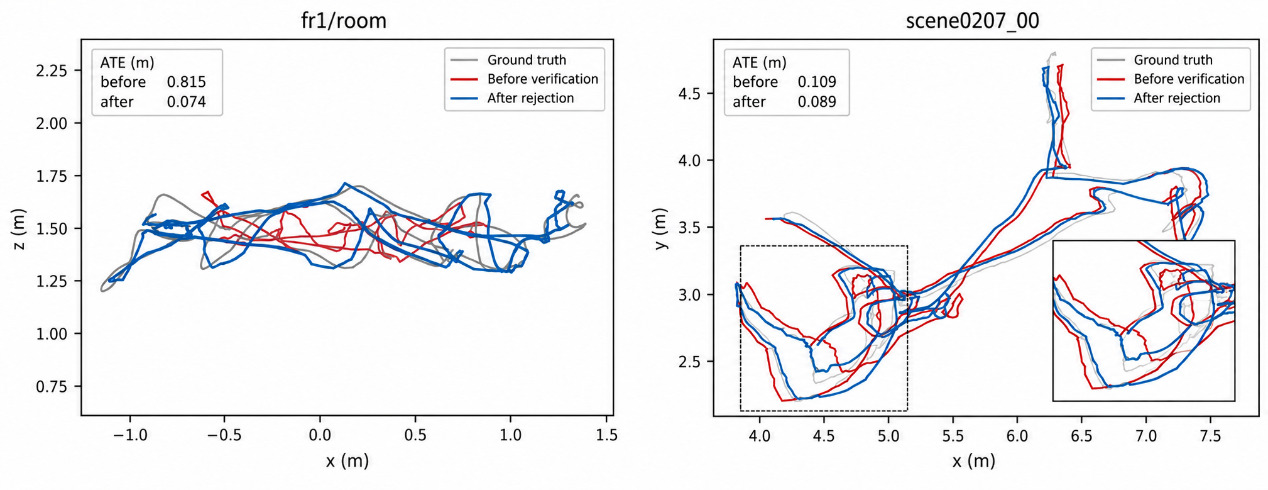}
\caption{Trajectories before and after loop verification: (a) fr1/room; (b) ScanNet scene0207\_00. Gray: ground truth; red: optimized trajectory without removing false loops; blue: trajectory with the proposed loop verification.}
\label{fig:traj}
\end{figure*}
\begin{table}[!t]
\caption{Loop verification and trajectory results on long ScanNet sequences}
\label{tab:3}
\centering
\footnotesize
\setlength{\tabcolsep}{2.0pt}
\renewcommand{\arraystretch}{1.07}
\begin{tabular}{@{}lccccc@{}}
\toprule
\textbf{Scene} & \textbf{Frames} & \textbf{Loops} & \textbf{False} & \textbf{ATE$_b$} & \textbf{ATE$_a$} \\
\midrule
scene0059\_00 & 1787 & 7 & 0 & 0.091 & 0.091 \\
scene0000\_00 & 5578 & 12 & 0 & 0.140 & 0.140 \\
scene0106\_00 & 2324 & 4 & 0 & 0.121 & 0.121 \\
scene0207\_00 & 1988 & 8 & 2 & 0.109$\times$ & 0.089 \\
\bottomrule
\end{tabular}
\par\vspace{2pt}\raggedright\footnotesize \emph{Note:} $\mathrm{ATE}_b$ and $\mathrm{ATE}_a$ denote absolute trajectory error (m) before and after loop verification, respectively. Before verification, all SALAD candidates are accepted; $\times$ marks false-loop-induced drift. Lower is better.
\end{table}

As shown in Table III, when all candidates are true, the verification module retains all valid geometric constraints and the trajectory accuracy is unchanged. Only when false loops are detected does the module remove erroneous constraints and correct the back-end optimization. On scene0207\_00, the method identifies and removes two false loops, reducing ATE from 0.109 m to 0.089 m. For the other scenes, which have no false candidates, the method retains valid loop constraints without reducing pose accuracy.

The trajectory correction achieved by loop verification is shown in Fig. 5. False constraints introduce mutually incompatible geometric relations in the pose graph and pull estimated poses away from ground truth. Together with the ATE values in Tables II and III, the trajectory plots show that filtering false-positive loops removes conflicting constraints and improves agreement with ground truth.

\subsection{Geometry-Anchored Persistent Instance Mapping}

\begin{figure*}[!b]
\centering
\includegraphics[width=0.90\textwidth]{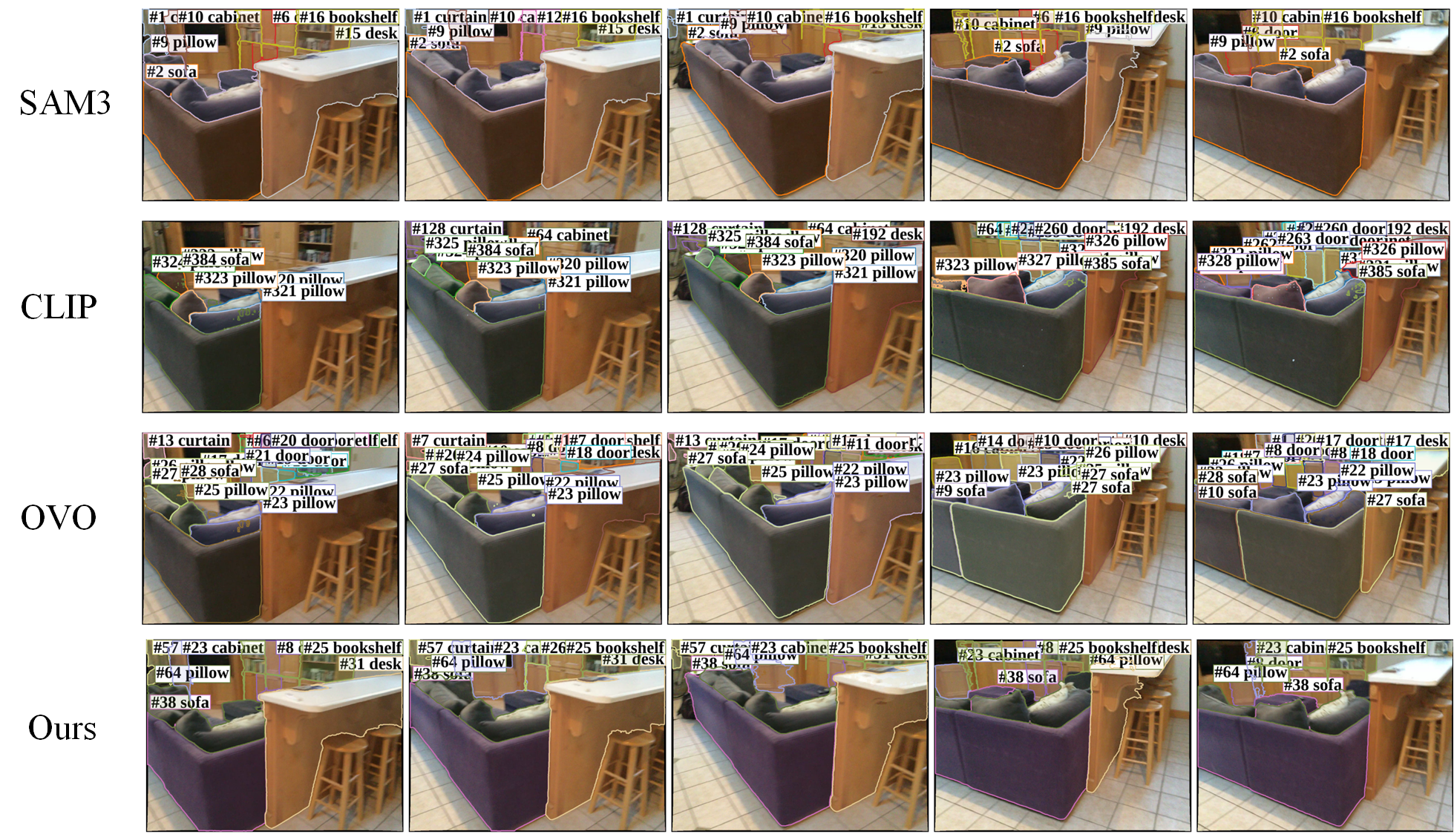}
\caption{Cross-view visualization of persistent instance identities on long ScanNet sequences. The first three columns show consecutive keyframes and the last two show revisit frames. Different colors denote different global instance IDs. From top to bottom: SAM3 Tracking, CLIP Association, OVO, and V3DSAM.}
\label{fig:identity}
\end{figure*}
\begin{table*}[!t]
\caption{Persistent instance identity comparison on long ScanNet sequences}
\label{tab:4}
\centering
\footnotesize
\setlength{\tabcolsep}{9pt}
\renewcommand{\arraystretch}{1.13}
\begin{tabular}{@{}lrrrrrr@{}}
\toprule
\textbf{Method} & \textbf{IDC $\uparrow$} & \textbf{ReID-R $\uparrow$} & \textbf{Inst. Coverage $\uparrow$} & \textbf{MOTA $\uparrow$} & \textbf{4D-PQ $\uparrow$} & \textbf{IDS $\downarrow$} \\
\midrule
SAM3 Tracking & 0.146 & 0.000 & 0.223 & \textbf{-0.184} & 0.118 & \textbf{285} \\
CLIP Association & 0.189 & 0.058 & 0.594 & -6.671 & \textbf{0.244} & 3263 \\
OVO (Vp+Vd) & 0.439 & 0.183 & \textbf{0.975} & -13.019 & 0.026 & 2994 \\
V3DSAM (Ours) & \textbf{0.565} & \textbf{0.192} & 0.736 & -0.286 & 0.110 & 873 \\
\midrule
OVO (GTp+GTd) & 0.631 & 0.317 & \textbf{0.918} & -4.603 & 0.046 & 1488 \\
V3DSAM (GTp+GTd) & \textbf{0.745} & \textbf{0.481} & 0.903 & \textbf{-2.805} & \textbf{0.054} & \textbf{901} \\
\bottomrule
\end{tabular}
\par\vspace{2pt}\raggedright\footnotesize \emph{Note:} Vp/Vd: VGGT-estimated pose/depth; GTp/GTd: ground-truth pose/depth. \textbf{Bold} denotes the best result within each input setting; the non-GT (first four) and GT (last two) rows are ranked separately. $\uparrow$/$\downarrow$ indicate preferred directions.
\end{table*}

Table IV compares instance identity consistency on ScanNet with SAM3, CLIP, and OVO baselines \cite{r4}, \cite{r22}, \cite{r23}. OVO is evaluated using both ground-truth poses and depth (GTp+GTd) and poses and depth predicted by the VGGT front end (Vp+Vd).

Among the external baselines, SAM3 Tracking, which relies entirely on temporal mask propagation, has low long-term reidentification performance under our test configuration. Its ReID-R of zero indicates that none of the evaluated instances retained a matching identity across the designated revisits. CLIP Association, which relies on 2D appearance features, also shows weak revisit identity scores and a large number of ID switches. These results illustrate the difficulty of maintaining global identity under viewpoint and occlusion changes using only temporal 2D tracking or appearance.

Two conclusions follow from comparing V3DSAM under different geometric inputs. First, feed-forward geometry quality substantially affects instance identity consistency. Replacing the predicted geometry with ground truth markedly improves the targeted identity metrics, indicating that geometric noise is a major bottleneck for persistent instance mapping. Second, when both methods use the same ground-truth geometry, V3DSAM outperforms OVO (GTp+GTd) on IDC and ReID-R, as well as MOTA and 4D-PQ. This ablation shows that OVO benefits substantially from ground-truth input geometry, while, after controlling for input quality, our geometry-anchored partitioning method exhibits stronger cross-view instance association.

With ground-truth geometry, V3DSAM's IDC and ReID-R increase substantially, whereas MOTA and 4D-PQ decrease. This indicates a complex relationship among geometric input quality, instance partitioning and matching, and composite metrics. More complete depth observations may introduce additional predicted regions, affecting false-positive and segmentation matching statistics. Multiple metrics must therefore be considered together when evaluating instance identity quality.

To isolate the effect of the class constraint, we remove the requirement of semantic-class agreement from the 3D adjacency graph while retaining all other mapping parameters. Across four ScanNet scenes, removing the class constraint reduces IDC from 0.565 to 0.220, instance coverage from 0.736 to 0.236, and 4D-PQ from 0.110 to 0.018. Geometric connected components alone can incorrectly merge nearby objects of different categories, causing severe under-segmentation. These results demonstrate that the class constraint is important for preserving the discriminability of global instances.

\Needspace{7\baselineskip}
Fig.~6 compares persistent instance identities across consecutive keyframes and long-term revisits on ScanNet. V3DSAM retains the same physical entities' identity labels more consistently across revisits, while the three baselines exhibit varying degrees of ID switches, incorrect matching, or excessive instance fragmentation. Together with Table IV, the results indicate that OVO obtains relatively high instance coverage but weaker long-term identity metrics; higher detection coverage does not necessarily imply more consistent cross-view instance identity.

\section{Conclusion}
This paper presents V3DSAM, an open-vocabulary semantic mapping system for feed-forward dense SLAM with two complementary association mechanisms: semantic-geometric dual-scale loop verification and geometry-anchored persistent instance mapping. The first mechanism identifies false-positive loops using only explicit geometry and semantic outputs, without depending on backbone-internal features. The second assigns global instance IDs to class-constrained 3D connected components and avoids inter-frame temporal ID propagation. Experiments on TUM RGB-D and ScanNet demonstrate that V3DSAM suppresses trajectory drift from false loops, achieves localization accuracy comparable to native VGGT verification, and improves long-term instance identity consistency relative to the evaluated baselines.

\end{document}